\documentclass{article}

\usepackage{PRIMEarxiv}

\usepackage[utf8]{inputenc} 
\usepackage[T1]{fontenc}    
\usepackage{hyperref}       
\usepackage{url}            
\usepackage{booktabs}       
\usepackage{multirow}%
\usepackage{amsfonts}       
\usepackage{nicefrac}       
\usepackage{microtype}      
\usepackage{lipsum}
\usepackage{fancyhdr}       
\usepackage{graphicx}       
\usepackage{amsmath,amssymb,amsfonts}%
\usepackage{amsthm}%
\usepackage{mathrsfs}%
\usepackage[title]{appendix}%
\usepackage{xcolor}%
\usepackage{textcomp}%
\usepackage{manyfoot}%
\usepackage{algorithm}%
\usepackage{algorithmicx}%
\usepackage{algpseudocode}%
\usepackage{listings}%
\usepackage{tabulary}
\usepackage{tabularx,booktabs}
\usepackage{geometry}
\usepackage{array}
\usepackage{lmodern}

\title{Design and Validation of a Responsible Artificial Intelligence-based System for the Referral of Diabetic Retinopathy Patients
\thanks{Under Review}}

\author{
  E. Ulises Moya-Sánchez\thanks{Also with Instituto Tecnológico  José Mario Molina Pasquel   Y Henríquez / TECMM}, Abraham Sánchez-Perez, Raúl Nanclares Da Veiga,  \\ \textbf{Alejandro Zarate-Macías, Edgar Villareal,  Alejandro Sánchez-Montes\thanks{Also with Secretaria de Salud Jalisco}}, \\ \textbf{Edtna Jauregui-Ulloa\thanks{Also with Universidad de Guadalajara}, Héctor Moreno} \\
  Gobierno de Jalisco,   \\
  Guadalajara Jalisco, Mexico\\
  \texttt{ulises.moya@iieg.gob.mx} \\
   \And
Ulises Cortés \\
Universitat Politècnica de Catalunya, Barcelona Supercoputing Center\\
Barcelona, Spain.\\
}

\begin{document}
\maketitle

\begin{abstract}
Diabetic Retinopathy (DR) is a leading cause of vision loss in working-age individuals. Early detection of DR can reduce the risk of vision loss by up to 95\%, but a shortage of retinologists and challenges in timely examination complicate detection. Artificial Intelligence (AI) models using retinal fundus photographs (RFPs) offer a promising solution. However, adoption in clinical settings is hindered by low-quality data and biases that may lead AI systems to learn unintended features. To address these challenges, we developed RAIS-DR, a Responsible AI System for DR screening that incorporates ethical principles across the AI lifecycle. RAIS-DR integrates efficient convolutional models for preprocessing, quality assessment, and three specialized DR classification models. We evaluated RAIS-DR against the FDA-approved EyeArt system on a local dataset of 1,046 patients, unseen by both systems. RAIS-DR demonstrated significant improvements, with F1 scores increasing by 5-12\%, accuracy by 6-19\%, and specificity by 10-20\%. Additionally, fairness metrics such as Disparate Impact and Equal Opportunity Difference indicated equitable performance across demographic subgroups, underscoring RAIS-DR's potential to reduce healthcare disparities. These results highlight RAIS-DR as a robust and ethically aligned solution for DR screening in clinical settings. The code, weights  of RAIS-DR are available at \url{https://gitlab.com/inteligencia-gubernamental-jalisco/jalisco-retinopathy} with RAIL. 
\end{abstract}


\section{Introduction}\label{sec:intro}

The prevalence of Diabetes Mellitus (DM) has been steadily increasing globally, with a particularly pronounced escalation in low and middle-income countries like Mexico \cite{WorldHealthOrganization, basto2023prevalencia}. 
Moreover, there is a strong correlation between the population size of DM patients and Diabetic Retinopathy (DR) patients~\cite{cheloni2019global} indicating a need for specialists, resources, and screening programs for DR early detection~\cite{kropp2023diabetic}.


One solution is to use AI-based systems to help grade DR specifically by classifying Retinal Fundus Photographs (RFP) for DR screening~\cite{arenas2022clinical}. AI-based DR graduation offers several potential benefits, such as increasing efficiency by maximizing clinical coverage, scaling the screening program, and helping to provide early DR detection and treatment. 
However, numerous examples demonstrated that AI health systems perform exceptionally well with specific curated datasets~\cite{gulshan2016development,willemink2020preparing,Topol2019}, but their performance is often limited when deployed in clinical environments~\cite{beede2020human,roberts2021common}. 
The main reasons for the limited performance in real-world applications are: i) the presence of low-quality data~\cite{beede2020human}, ii) the model learning something unexpected or spurious features such as the background~\cite{degrave2021ai}, and iii) the lack of large, representative labeled datasets~\cite{willemink2020preparing}. This work introduces a novel redesigned Responsible AI System for DR Screening (RAIS-DR),  which integrates ethical principles into the classic AI lifecycle to address these challenges.

The responsible principles underpinning RAIS-DR are derived from a consensus mapping of ethical principles~\cite{fjeld2020principled}. Transforming these principles into effective and practical actions/operations is neither trivial nor a simple task that can be automated. This strategy involved new responsible stages/actions in the typical AI lifecycle. See Table \ref{tab:responsibleAI} in Section \ref{sec:RAI} for the link between these stages, the ethical principles, and the AI lifecycle stages.


RAIS-DR was redesigned with six efficient Convolutional Neural Networks (ConvNets), incorporating workflow steps to assess image quality in real-time and enable retaking of RFPs if necessary. Each ConvNet was trained, fine-tuned, and tested using publicly available datasets, including EyePACS~\cite{KaggleCompetition}, IDRiD~\cite{idrid}, and MESSIDOR II~\cite{messidor}. As noted by \cite{roberts2021common}, one of the most common pitfalls in real-world AI systems is testing only on the test split from the same dataset. To address this, the models underwent retrospective clinical validation using a separate local dataset in a pilot study involving 1,046 patients, with ground truth labels determined by consensus among three retina experts. Performance was evaluated using F1-score, sensitivity, specificity, Positive Predictive Value (PPV), Negative Predictive Value (NPV), and accuracy.

The experimental setup focused on detecting Referable DR (RDR) and All-Cause Reference (ACR) cases. RDR was defined as severe non-proliferative or proliferative DR according to Mexican clinical guidelines~\cite{imss2015guiadet}, while ACR included RDR cases and ungradable images. This approach ensures that RAIS-DR aligns with clinical standards and addresses real-world diagnostic challenges.


Another key contribution of this paper is the comparative evaluation of two AI screening systems, including a fairness assessment. We compared RAIS-DR with the FDA-approved EyeArt system (Eyenuk, \url{https://www.eyenuk.com}) on both a per-patient and per-image basis. RAIS-DR outperformed EyeArt, achieving an accuracy improvement of up to 19\%. To evaluate fairness, we assessed potential biases using Disparate Impact (DI) and Equal Opportunity Difference (EOD) across factors such as sex, projection type (macula-centered vs. optic nerve-centered), laterality (left vs. right eye), and age. DI measures the ratio of positive outcomes between subgroups, with values close to 1 indicating fairness, while EOD evaluates the difference in true positive rates between subgroups, with values close to zero suggesting equitable performance. The DI values ranged from 0.984 to 1.031, indicating minimal bias and balanced treatment across subgroups. Additionally, EOD values were close to zero, further confirming RAIS-DR's fairness across all evaluated subgroups.

GradCAM~\cite{selvaraju2017grad} was also used to analyze and visualize the inference gradients of the primary (referral) model. Finally,  a git repository with cloud deployment facilitates evaluation of the RAIS-RD  by other retina experts.

The remainder of the paper is organized as follows: Section \ref{sec:relatedwork} presents the literature review. Section \ref{sec:RAI} describes the RAIS-DR  and the implementation of ethical principles across the AI lifecycle. Section \ref{sec:data} details the training data and the local clinical dataset used for validation. Section \ref{sec:experimental} outlines the experimental setup and evaluation metrics. Finally, the results and conclusions are presented in Section \ref{sec:results} and Section \ref{sec:conlusions}, respectively.

\section{Literature review}\label{sec:relatedwork}

Previous work \cite{gulshan2016development}, developed and validated a DL model on two external datasets (EyePACS and Messidor II), achieving $>90\%$ sensitivity and specificity in distinguishing Referable-DR (Moderate DR or worse per ICO grades) from Non-Referable DR. However, Beede \textit{et al}.\cite{beede2020human} reported that, when deployed in Thai clinics, it reduced false negatives by 23\% reduction while increasing false positives by 2\%. These performance issues have been attributed primarily to low-quality images, which impact gradability, a common challenge in DR screening due to other ocular conditions such as cataracts~\cite{arenas2022clinical,pinedo2022suitability,lin2020retinal}. To address this,  quality assessment models were integrated into the AI system presented here to prompt retakes of RFPs or facilitate direct patient referrals when necessary.

In \cite{korot2021code}, code-free cloud tools were used to train a DR classification model with minimal preprocessing, resizing images to 224 $\times$ 224 pixels. However, external validation showed highly variable classification accuracy for Referable DR (RDR) versus Non-Referable DR (NRDR), dropping to 10-13\%. This variability could be attributed to factors such as input size, background noise in retinal fundus photographs (RFP), and geometric distortions from resizing, as identified in prior work
\cite{briceno2020automatic}. To address these issues, the proposed approach preprocesses RFPs, increases ConvNet input size to 512 $\times$ 512 pixels, and includes steps to remove background noise, spurious features such as letters, and geometric distortions, thereby enhancing model robustness~\cite{ribeiro2016why,dodge2016understanding}

Scheetz et. al.~\cite{scheetz2021real} presented an AI-assisted DR screening with 236 participants in Australia, achieving an AUC, sensitivity, and specificity of 0.92\%, 96.9\%, and 87.7\%, respectively. In addition, the study highlighted the most important challenges, image quality, data flow, and technology adoption by clinical users. A key distinction in the presented work is the use of a larger patient dataset (1046), combined with post-training analysis of the model's explainability and bias measurement. This study also compares RAIS-DR system performance with an external AI screening system (Eyeart).

AI system failures in healthcare are common~\cite{roberts2021common,degrave2021ai}, frequently stemming from issues such as insufficient external validation, unreported limitations, model bias, and lack of explainability. One approach to address these challenges is to apply a responsible AI framework. However, existing frameworks, tools, and checklists for responsible AI~\cite{mongan2020checklist, wolff2019probast} are often too general,  and static, lacking adaptability to specific real-world healthcare needs\cite{chomutare2022artificial}. These tools also offer limited feedback and fail to address challenges such as the use of non-tabular data, clinical implications, and technical constraints. An iterative process, like the AI lifecycle, enables early identification and resolution of these issues, reducing failure risks and enabling regular updates based on feedback. To ensure responsible development,
the RAIS-DR system includes additional stages in the AI lifecycle;  see Table \ref{tab:responsibleAI} for details.

\section{Responsible AI System}\label{sec:RAI}

A key objective of responsible AI systems is to address ethical, legal, technical, and societal issues in real-world applications a priority for governments, businesses, and society. 
However, responsible AI is complex and interdisciplinary, so reducing it to a checklist risks neglecting ongoing monitoring and prioritizing short-term gains over long-term ethics and safety. To mitigate these issues, iterative responsible actions are implemented throughout the AI lifecycle~\cite{advanceaccountabilityai2023}.


From a large set of responsible and ethical principles~\cite{fjeld2020principled,advanceaccountabilityai2023}. This work focuses on the following:  professional responsibility, privacy, accountability, safety, security, transparency, explainability, human control of technology, fairness and non-discrimination, and promotion of human values and well-being. 
Furthermore, the responsible AI system pipeline was developed following the local clinical guidelines~\cite{imss2015guiadet} and accounts for model limitations.

These responsible stages are outlined in Table \ref{tab:responsibleAI}, along with their corresponding implemented actions, underlying principles, and related AI lifecycle stages~\cite{advanceaccountabilityai2023}. Note that each responsible stage may correspond to more than one responsible action.  An important outcome of incorporating responsible AI practices was redesigning the system pipeline, focusing on data exploration, analysis, and standardization. Additionally,  fairness metrics and explainability were implemented as well. Benchmarking and exposure to external review were crucial in addressing ethical, legal, technical, and societal considerations, ensuring a more comprehensive and accountable approach to AI-based technology development.

\begin{figure}[h!]
\centering
\includegraphics[scale=0.35]{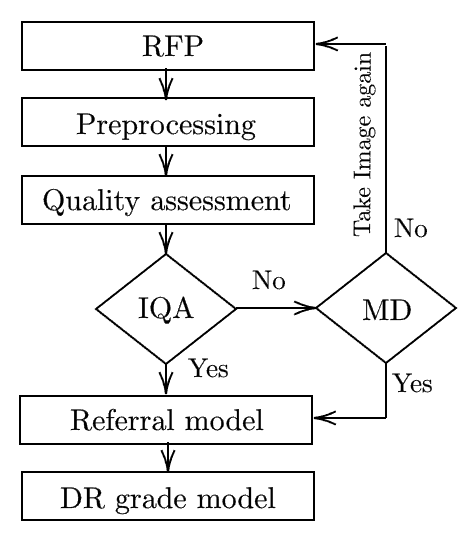}
\caption{The RFP is the system input, which undergoes preprocessing to remove the background. Then, quality models evaluate the image in general, and an additional model detects the macula and the optic nerve. If the image is not gradable (low quality), the  IQA=Image Quality Assessment and an MD=Medical Decision define if the image should be retaken. }
\label{fig:models2}
\end{figure}

Figure~\ref{fig:models2} illustrates the RAIS-DR pipeline. The process begins with an RFP acquisition. The background is removed in the \textit{Processing} stage, and the image is adjusted to the appropriate aspect ratio and size. Subsequently, the\textit{ Quality Assessment} stage determines whether the image is of sufficient quality to be graded. Based on the quality assessment results, the observed anatomy, and potential presence of other eye conditions such as cataracts, a trained professional must make an on-site medical decision (MD) regarding whether to retake the RFP. Following this, three convolutional networks are used, one for patient referral classification and two for grading Diabetic Retinopathy (DR) severity according to International Council of Ophthalmology (ICO) standards. These DR grading models are essential for adhering to clinical guidelines, providing clear results to physicians and retina specialists, and supporting prioritization in patient care.

\begin{table}[ht!]
    \centering
\caption{AI-Responsible stages, actions, and principles implemented in the AI lifecycle. }
    \begin{tabular}{>{}p{1.8cm}>{}p{4.5cm}>{}p{1.5cm}>{}p{2.9cm}}
     \toprule
\textbf{AI Responsible stage} & \textbf{ Responsible actions} & \textbf{AI lifecycle stages} & \textbf{Responsible principles}  \\
\midrule
       { Ethical and technical review}  & Conducted iterative reviews to ensure benefit to society, ethical principles, and technical robustness.  & Plan and design. &  Professional responsibility, privacy, safety and security, promotion of human values and well-being. \\
         & Ensured control over data use, privacy, consent, ability to restrict processing, and the rights to rectification and erasure.  & Collect and process data. &   \\ 
          & Addressed dispute resolution and environmental responsibility. & Operate and monitor.  &   \\
          \midrule
System redesign & Developed custom models: created new models for image quality assessment. & Build and use model. & Accountability, Human control of technology. \\
 & Adapted a custom data pipeline according to clinical guidelines and model limitations. &  Deploy. &\\        
          \midrule
Exploration, analysis, standardizing data         & Avoided spurious features and applied preprocessing to maintain aspect ratio. & Collect and process data. & Robustness, Professional responsibility.\\
\midrule
Bias analysis, explainability & Incorporated fairness metrics into the model evaluation. & Verify and validate. &  Fairness and nondiscrimination. \\
 & Computed gradient maps to highlight the learned features. &  &  Explainability. \\
\midrule
Monitoring & Deploying a cloud system for feedback from retina experts. & Deploy, Verify, and validate. & Robustness, human control of technology.\\
\midrule
Benchmarking and exposure to external review & Methods and results to peer-review &  & Robustness, and transparency. \\
 & Compared performance versus FDA-approved system. & &  \\
 & Published models and responsible AI practices. & & \\

          \bottomrule

    \end{tabular}
    \label{tab:responsibleAI}
\end{table}

Table \ref{tab:rais-dr-models} summarizes the RAIS-DR models, classes, ConvNet backbones, and the main actions. It is important to note that we tested with at least  five different models (including visual transformers) and best trade off between performance and inference time is presented at Table\ref{tab:rais-dr-models}.  The MP model removes the background, while the MQ and MA are models to define quality assessment and the anatomical detection of optic nerve and macula. Model M1-Referral is the most important model for classifying if the patient is referred, and finally, M2-DR grade and M3-DR grade help us to do fine-grain DR-level classification. The entire process takes approximately two seconds to execute all models on a CPU cloud server.   

\begin{table}[ht!]
\caption{Summary of RAIS-DR Models, backbones, and main actions. }
        \label{tab:rais-dr-models}
\centering
    \begin{tabular}{clll}
    \toprule
    \textbf{Model} & \multicolumn{1}{c}{\textbf{Classes}} & \textbf{Backbone} & \textbf{Action} \\
    \midrule
    \multirow{2}{*}{MP} & 0: Background & \multirow{2}{*}{Unet \cite{ronneberger2015u}} & \multirow{2}{*}{Remove Background}  \\
     & 1: Eye &  &  \\
      \midrule
     \multirow{2}{*}{MQ} & 0: Bad quality & \multirow{2}{*}{MobileNet \cite{howard2017mobilenets}} & \multirow{2}{*}{Evaluate Quality}  \\
     & 1: God quality &   &  \\
     \midrule
      \multirow{2}{*}{MA} & 0: Macula detection & \multirow{2}{*}{Faster R-CNN \cite{ren2015faster}} & \multirow{1}{*}{Anatomical detection}  \\
     & 1: Optic nerve detection &  & Repeat or M1  \\
      \midrule
    \multirow{2}{*}{M1-Referral} & 0: \{R0,R1,R2\} & \multirow{2}{*}{EfficientNet V2-s \cite{tan2104efficientnetv2}} & Evaluate M2  \\
     & 1: \{R3,R4\} &   & Evaluate M3 \\
    \midrule
    \multirow{2}{*}{M2-DR grade} & 0: \{R0,R1\} & \multirow{2}{*}{EfficientNet V2-s \cite{tan2104efficientnetv2}} & Primary care review in 12 months \\
     & 1: \{R2\} &  & Primary care review in 6 months  \\
    \midrule
    \multirow{2}{*}{M3-DR grade} & 0: \{R3\} & \multirow{2}{*}{EfficientNet V2-s \cite{tan2104efficientnetv2}} & \multirow{2}{*}{Referral to retina specialist. } \\
     & 1: \{R4\} &  & \\
    \bottomrule
    \end{tabular}
\end{table}

\section{Data and labels}\label{sec:data}

All the RFP were labeled according to the ICO severity scale: no apparent DR (R0), mild non-proliferative DR (R1), moderate non-proliferative DR (R2), severe non-proliferative DR (R3), proliferative DR (R4). In addition, for the local data, (R5) is defined as eye enucleation, and (R6) is assigned to ungradable due to inadequate image quality to confirm the grade of DR. 

\subsection{Public data: for training}

Three public datasets were used to train our models:  EyePACS \cite{KaggleCompetition}, IDRiD~\cite{idrid}, and MESSIDOR II~\cite{messidor}. Most of the RFP in these datasets are 45-degree field, macula centered, with different spatial resolutions, aspect ratios, cameras, and image quality. No additional demographic information is available for these public datasets.

\subsection{Local data: only for  external clinical validation}
A local dataset was gathered from three primary care facilities in Jalisco, \texttt{La Aurora y la Esperanza, Yugoslavia, and  Paraísos del Colli} with the support from the IDB Fair-LAC program between August 2021 and March 2022. These clinics use three Topcon non-mydriatric cameras TRC-NW400.

The inclusion criteria for sample selection included patients with  DM type II, both male, and female aged eighteen years and above, from the Cardio-metabólicas program. Two RFP   one macula-centered and one optic nerve-centered were captured per eye.   Mydriasis was used during this RFP acquisition for at least 498 patients. For more information about the data collected see~\cite{tejerina2023implementacion}. 
The ethical protocol was registered with the ID number \texttt{9/E-JAL/2021} by the Secretaria de Salud de Jalisco. 
Patients choosing to participate were required to sign an informed consent. This study relied exclusively on anonymous data. The RFP was labeled by three selected Colegio de Oftalmológos de Jalisco members. Table \ref{tab:localdata_characteristics} presents the main local data characteristics,  demographics, image quality distribution, and disease severity distribution according to ICO.


\begin{table}[h]
\centering
\caption{Summary of data characteristics, demographic information, image quality distribution, and  DR grade distribution of the local data for external validation.} \label{tab:localdata_characteristics}
\begin{tabular}{ll}
\toprule
\textbf{Characteristics} & \multicolumn{1}{l}{\textbf{Local Data Set}} \\
\midrule
No. of images & 4296 \\
No. of  optic nerve-centered images  & 2148 \\

No. of  macula-centered images  & 2148 \\
No. of ophthalmologists & 3 \\
No. of grades per image & 3 \\
\midrule
\multicolumn{2}{l}{Patient demographics} \\
\midrule
\quad No. of unique individuals & 1074 \\
\quad Age, mean (SD) &  60.4 (12.1) \\
\quad \begin{tabular}[c]{@{}l@{}}Female, No./total (\%) among patients \\ for which sex was know \end{tabular} & 688/1046 (65.7)  \\
\midrule
\multicolumn{2}{l}{Image quality distribution} \\
\midrule
\quad \begin{tabular}[c]{@{}l@{}}Fully gradable, No./total (\%) among patients\\ for which image quality was assessed\end{tabular} & 797/1046 (76.2) \\
\quad No image (patients) & 28 \\
\midrule
\multicolumn{2}{l}{\begin{tabular}[c]{@{}l@{}}Disease severity distribution classified by \\ majority decision of ophthalmologists \\ (reference standard)\end{tabular}} \\
\midrule
\quad \begin{tabular}[c]{@{}l@{}}Total patients for which both diabetic\\ retinopathy and diabetic macular edema \\ were assessed, No. (\%)\end{tabular} & 797 (100) \\
\quad R0. No diabetic retinopathy & \multirow{2}{*}{679} \\
\quad R1. Mild diabetic retinopathy &  \\
\quad R2. Moderate diabetic retinopathy & 64 \\
\quad R3. Severe diabetic retinopathy & 28 \\
\quad R4. Proliferative diabetic retinopathy & 26 \\
\quad Referable diabetic macular edema & 0 \\
\quad (R3+R4) Referable diabetic retinopathy & 54\\
\bottomrule
\end{tabular}
\end{table}


\section{Experimental setup}\label{sec:experimental}

All the models were trained with data augmentation (horizontal mirroring, rotation, blur,  class balancing, and transfer learning (ImageNet pre-trained model). 
The most relevant modification in the model architecture is the input shape (512x512x3) and the last layer to make a binary model. We use an IBM AC922 and an Nvidia DGX A-100 640 server to train and fine-tune the models and the PyTorch framework to implement the models.

We evaluated different clinical scenarios. First, we assessed the performance of Referable Diabetic Retinopathy (RDR) detection per patient, excluding images that were not fully gradable. According to the Mexican DR guidelines, an RDR case is defined as either severe non-proliferative DR (R3) or proliferative DR (R4)~\cite{imss2015guiadet}. Next, we presented the All-Cause Referable Cases (ACR) classification results per patient (including images that were not fully gradable).  ACR is defined as severe non-proliferative DR (R3), proliferative DR (R4), or ungradable image  (R6). Performance per image was evaluated for both RDR and ACR classifications, comparing it with the FDA-approved model (Eyeart) in each case.

System's performance was further examined for bias across different subgroups, including sex, eye projection, age, and laterality. The Disparate Impact $DI$, was used to compare the favorable outcomes for a monitored group against those for a reference group.  In  $DI$ interpretation, usually when  $1<DI<0.8$ (known as the \textit{four-fifths rule}), the model does not have a bias for a group. 
Equal Opportunity Difference ($EOD$) quantifies the deviation from equality of opportunity, ensuring that the same proportion of favorable results is reached across groups. Ideally, this metric must be equal zero. 


Table \ref{tab:experimental_setup} the experimental setup cases. The first two experiments simulate a clinical environment, focusing on patient referral classification. Experiments 3 and 4 assess the model's referral performance on individual images. Finally, experiments 5 through 9 are designed to evaluate potential biases in the model. The abbreviations used in Table \ref{tab:experimental_setup} are defined as following: Referable Diabetic Retinopathy (RDR) Class 0 as "Not referral" (R0,R1,R2) and class 1 as DR referral (R3,R4). All-cause referable (ACR) defines Class 0 as "Not referral" (R0,R1,R2) and Class 1 as "referral" (R3,R4,R6). Moreover, the projection centered on the macula is defined in the table as Proj A. The projection centered on the optic nerve is defined as B. "Both" refers to both sex: male (M) and Female (F) patients. "L" is  Left , and "R" is  Right eye.

\begin{table}[ht]
    \centering
        \caption{Summary of the experimental setup cases. See the text paragraph for the abbreviation definitions.}
    \begin{tabular}{lllllll}
    \toprule
      No &  Data &    Type & Proj & Sex & Laterality & Age \\
    \midrule
     1 &{RDR}    & { Patient} & {A} & {Both} & {All} & {All} \\
                         
    2 & {ACR} & { Patient} & {A} & {Both} & {All} & {All}  \\
    3 & {RDR}  & {Per image} & {A} & {Both} & {All} & {All} \\                       
     4 & {ACR}  & {Per image} & {A} & {Both} & {All} & {All} \\
                            
    5 & {RDR}  & {Per image} & { A Vs B} & {Both} & {All} & {All} \\
                            
   6 & {RDR}  & {Per image} & { A Vs AB} & {Both} & {All} & {All} \\
                            
   7 & {RDR}  & {Patient} & { A } & {M Vs F} & {All} & {All} \\
                            
   8 & {RDR}  & {Per image} & { A } & {Both} & {L Vs R} & {All} \\
                            
   9 & {RDR}  & {Patient} & { A } & {Both} & {All} & { $\le$60 } \\

    \bottomrule                       
    \end{tabular}

    \label{tab:experimental_setup}
\end{table}


The metrics used for the classification evaluation are based on the F1 score:(Sensitivity, Specificity, Positive Predictive Values, Negative Predicted Values, and Accuracy).  In addition, the Receiver-operating characteristic curve (ROC) is employed as a visual representation of model performance across all model thresholds. Moreover,  the Area Under the Curve (AUC) is measured for each ROC curve. In the following equations, we define the equation of each metric~\cite{caton2024fairness} : 

\begin{eqnarray}
     Accuracy &=& \frac{TP+TN}{TP+TN+FP+FN}, \label{eq:acc} \\
     PPV\, (\text{precision}) &=& \frac{TP}{TP+FP}, \\
     NPV &=& \frac{TN}{TN+FN}, \\
     Sensitivity\, (\text{recall}) &=& \frac{TP}{TP+FN}, \\
     Specificity &=& \frac{TN}{TN + FP} \\
     F1 &=& 2 * \frac{\text{precision} * \text{recall}} {\text{precision} + \text{recall}} \label{eq:f1} \\
\end{eqnarray}
where,  $TP =$ True Positive, $TN =$ True Negative, $FP =$ False Negative, and $FP = $ False Positive. The following equations~\cite{caton2024fairness} are used for the fairness evaluation:
\begin{equation}
    DI = \frac{P(R=1|A=\text{unprivileged})} {P(R=1|A=\text{privileged})}, 
\end{equation}

\begin{eqnarray}
    \begin{split}
    EOD_1 &= P(R=1 \mid Y=1, A=\text{unprivileged}) \\
          &\quad - P(R=1 \mid Y=1, A=\text{privileged}),
\end{split}
     \\
     \begin{split}
     EOD_0 &= P(R=0\mid Y=0, A=\text{unprivileged}) \\
     &\quad - P(R=0\mid Y=0,  A=\text{privileged}), 
\end{split}      
\end{eqnarray}
where unprivileged is the monitor group, privileged is the reference group and  $P(R=outcome|Y=label, A=group)$ represents the probability $P$ to get an outcome $R$ for a label $Y$  and a group $A$.  In this work $EOD_1$ when $Y=1$ and $EOD_0$ when $Y=0$.

To evaluate the explainability GradCam~\cite{selvaraju2017grad}   were used in all the RFP (using M1-Referral model). Explainability helps clinicians, creators,  patients, and regulatory bodies to understand the model errors. Note that we present only a few representative cases in this work.

\section{Results and analysis}\label{sec:results}
This section presents the results only using the local data (external clinical validation). 
\subsection{Patient referral classification}
 
\textit{Patient RD Reference (RDR)}. As depicted in Table \ref{tab:experimental_setup} experiment one, the images analyzed are macula-centered  (projection A), both sexes,  and the non-gradable images were removed. The dataset included 797 patients (743 classified as Non-Referable Diabetic Retinopathy (NRDR) and 54 as Referable Diabetic Retinopathy (RDR)). Table~\ref{tab:report_results_t1} presents a detailed classification report. The F1 score; (sensitivity, specificity, positive predictive value (PPV), negative predictive value (NPV), and accuracy) of the presented model here, is as follows: 98; (91, 96, 64, 99, 96). In comparison, the EyeArt system's performance is 86; (98, 76, 23, 100, 77). These results indicate that this model significantly outperforms EyeArt, particularly in specificity, PPV, F1 score, and overall accuracy, suggesting that the model effectively identifies RDR cases while maintaining a lower rate of false positives.








\textit{All-cause reference per patient}. This experiment is defined as number two  in Table \ref{tab:experimental_setup}. The most important difference from experiment one is that we include non-gradable images  (R6) in the referral class. Table~\ref{tab:report_results_t1} presents a classification report of All-Cause Reference (ACR) with 1046 patients (743 NRDR, 303 RDR). The AI systems demonstrated the following classification performance, 
F1;(sensitivity, specificity, PPV, NPV, accuracy): 
 RASIS-RD (Ours),  90;(89, 86, 72, 95, 87); 
 Eyeart            85;(93, 76, 61, 96, 81). RAIS-DR demonstrates superior performance in F1 score, specificity, PPV, and accuracy, exceeding EyeArt by 5\%, 10\%, 11\%, and 6\%, respectively.  Although EyeArt has a slight edge in sensitivity, our model has higher accuracy and PPV make it more robust in clinical applications where reducing false positives and maintaining overall accuracy are key for effective patient management.


%
%

\begin{table}[h]
\centering
\caption{Reference classification report per patient with local data, for both cases All-Cause Reference (ACR) and Reference DR RDR. The proposed system is the RASI-DR.}
\begin{tabular}{lllllllll}
\toprule
\textbf{DRD} & \multicolumn{8}{l}{} \\
\midrule
\textbf{Model} & \textbf{TN} & \textbf{FP} & \textbf{FN} & \textbf{TP} & \textbf{PPV(\%)} & \textbf{Sen(\%)} & \textbf{Spec(\%)} & \textbf{NPV(\%)} \\
\midrule
Proposed & \textbf{715} & \textbf{28} & 5 & 49 & \textbf{64} & 91 & \textbf{96} & 99 \\
Eyeart & 562 & 181 & \textbf{1} & \textbf{53} & 23 & \textbf{98} & 76 & \textbf{100} \\
\midrule
\multicolumn{9}{l}{\textbf{ACR}} \\
\midrule
\textbf{Model} & \textbf{TN} & \textbf{FP} & \textbf{FN} & \textbf{TP} & \textbf{PPV(\%)} & \textbf{Sen(\%)} & \textbf{Spe(\%)} & \textbf{NPV(\%)} \\
\midrule
Proposed & \textbf{637} & \textbf{106} & 33 & 270 & \textbf{72} & 89 & \textbf{86} & \textbf{95} \\
Eyeart & 562 & 181 & \textbf{21} & \textbf{282} & 61 & \textbf{93} & 76 & 96 \\
\bottomrule
\end{tabular}
    \label{tab:report_results_t1}
\end{table}

\begin{figure}[h]
    \centering
    \includegraphics[scale=0.47]{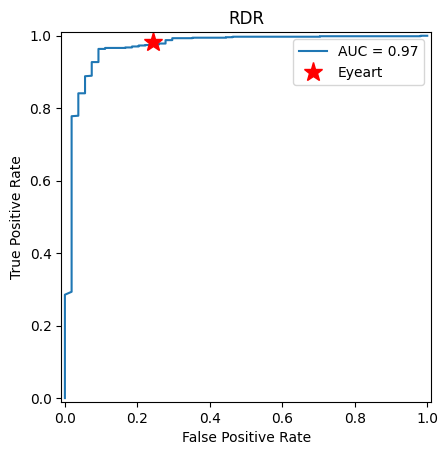}
     \includegraphics[scale=0.47]{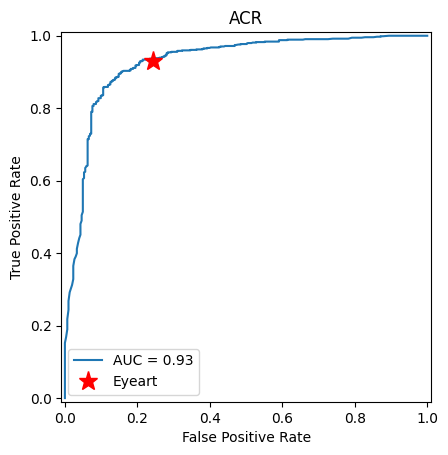}
    \caption{Left DRD ROC curve and the individual point of Eyeart system, right ACR ROC curve and the individual point of Eyeart system.} 
    \label{fig:roc_per_patient}
\end{figure}

Figure~\ref{fig:roc_per_patient} presents ROC curves and AUC values for our models, ACR (AUC=0.93) and DRD (AUC=0.97), using the external validation dataset. Additionally, individual points ({\color{red} \(\star\)}) represent the True Positive Rate (TPR) and False Positive Rate (FPR) of the EyeArt classification system. A key advantage of a locally trained model is the ability to access and adjust all relevant threshold values, providing control over the AUC. Clinical requirements exhibit significant variability across different medical applications, patient populations, and potential consequences of diagnostic errors. Therefore, maintaining precise control over AI-based systems is paramount for enhancing their efficacy, ensuring fairness, and optimizing patient outcomes.


\subsection{Image referral classification}

\textit{RDR per image} classification metrics are defined as follows F1;(sensitivity, specificity, PPV, NPV, accuracy): RAIS-DR,  98;(89, 96, 61, 99, 96);                                                        Eyeart 85;(98, 74, 19, 100, 75). The proposed model shows higher values for  F1 score, specificity, PPV, and accuracy,    by 13\%,   12\%, 42\%, and 21\%,  respectively. However, EyeArt has a higher sensitivity by 9\% and a slightly higher NPV 1\%. Table \ref{tab:report_results_per_image} lists the classification metrics per image for RDR for the proposed system and the Eyeart system. Overall, this performance profile suggests that the proposed model better suited for contexts where specificity and accurate classification of true negatives are prioritized.




 




\textit{ACR per image} classification following the same convention F1 score;(sensitivity, specificity, PPV, NPV, accuracy) our results are the following: Proposed,  92;(87, 89, 69, 96, 88); Eyeart 84;(93, 74, 51, 97, 78). The proposed model has higher values for  F1 score, specificity, PPV, and accuracy,    by 8\%,   15\%, 18\%, and 10\%,  respectively, and lower values for sensitivity by 6\% and NPV 1\%.  
Table \ref{tab:report_results_per_image} presents the classification report for ACR for both the proposed model and EyeArt.

\begin{table}[h]
\caption{Reference per image classification with ACR and DRD per image.}
\centering
\begin{tabular}{lllllllll}
\toprule
\textbf{RDR} & \multicolumn{8}{l}{} \\
\midrule
\textbf{Model} & \textbf{TN} & \textbf{FP} & \textbf{FN} & \textbf{TP} & \textbf{PPV(\%)} & \textbf{Sen(\%)} & \textbf{Spec(\%)} & \textbf{NPV(\%)} \\
\midrule
Proposed & \textbf{1550} & \textbf{58} & 11 & 89 & \textbf{61} & 89 & \textbf{96} & 99 \\
Eyeart & 1186 & 422 & \textbf{2} & \textbf{98} & 19 & \textbf{98} & 74 & \textbf{100} \\
\midrule
\multicolumn{9}{l}{\textbf{ACR}} \\
\midrule
\textbf{Model} & \textbf{TN} & \textbf{FP} & \textbf{FN} & \textbf{TP} & \textbf{PPV(\%)} & \textbf{Sen(\%)} & \textbf{Spe(\%)} & \textbf{NPV(\%)} \\
\midrule
Proposed & \textbf{1428} & \textbf{180} & 60 & 410 & \textbf{69} & 87 & \textbf{89} & {96} \\
Eyeart & 1186 & 422 & \textbf{33} & \textbf{437} & 51 & \textbf{93} & 74 & \textbf{97} \\
\bottomrule
\end{tabular}
    \label{tab:report_results_per_image}
\end{table}

The proposed classification model demonstrates strong performance and an excellent balance across various metrics, particularly excelling in F1 score, specificity, PPV, and overall accuracy, although it shows room for improvement in sensitivity and NPV. As mentioned, one way to improve sensitivity is to adjust the classification threshold or use data-centric techniques. 


\subsection{Fairness evaluation}\label{ssec:bias}
In the following section, we present the results of fairness metrics. 
In this case, we present the results with the local data per image regarding our AI-based system. 
As mentioned, we compare projections A (macula-centered) and B (optic nerve-centered), sex, laterality, and age (see Table \ref{tab:experimental_setup} experiments from five to nine). 
Table \ref{tab:report_bias} compares the $DI$, $EOD_1$, $EOD_0$. 
According to the results,  $0.984 < DI <1.031$, suggests the absence of adverse impact,  beyond the limitations of the four-fifths rule such as a small sample size, or a simplistic threshold~\cite{raghavan2024limitations}. Concerning the $EOD_0$ and $EOD_1$, the results are according to the DI; the EOD values are closer to zero \textit{i.e.} our models are fair with the different subgroups; in addition, the same subgroups have more significant values. These evaluations were important due to some convolutional models could predict the sex or age from an RFP~\cite{korot2021predicting,gerrits2020age}. In addition, it's important to examine other equity factors, such as long-term outcomes, or systemic barriers, that might still exist.

\begin{table}[h]
\caption{Fairness metrics, Disparate Impact (DI), and Equal Opportunity Difference  (EOD) when $Y=1$ and when $Y=0$. Both metrics suggest the absence of adverse impact.}
\centering
\begin{tabular}{lllllll}
\toprule
\multicolumn{6}{l}{\textbf{DRD}} \\
\midrule
\textbf{Feature} & \textbf{Type} & \textbf{unprivileged} & \textbf{privileged} & \textbf{$DI$} & \textbf{$EOD_0$} & \textbf{$EOD_1$} \\
\midrule
Projection & Per Image & B & A & 0.997 & -0.005 & -0.004 \\
Projection & Per Image & A & AB & 1.001 & 0.003 & 0.002 \\
Laterality & Per Image & Left & Right & 1.008 & 0.003 & 0.061 \\
Sex & Per Patient & Male & Female & 0.984 & -0.008 & -0.0204 \\
Age & Per Patient & $<$60 & $\ge$60 & 1.031 & -0.003 & 0.149 \\
\midrule

\multicolumn{6}{l}{\textbf{ACR}} \\
\midrule
\textbf{Featrure} & \textbf{Type} & \textbf{unprivileged} & \textbf{privileged} & \textbf{$DI$} & \textbf{$EOD_0$} & \textbf{$EOD_1$} \\
\midrule
Projection & Per Image & B & A & 1.00004 & -0.012 & 0.014\\
Projection & Per Image & A & AB & 0.999 & 0.006 & -0.007 \\
Laterality & Per Image & Right & Left & 0.9907 & 0.003 & -0.056 \\
Sex & Per Patient & Male & Female & 0.938 & -0.026 & -0.047 \\
Age & Per Patient & $<$60 & $\ge$60 & 1.066 & 0.001 & -0.047 \\
\bottomrule
\end{tabular}
  
    \label{tab:report_bias}
\end{table}

\subsection{Explainability}

\begin{figure}[ht]
    \centering
    \includegraphics[width=0.94\linewidth]{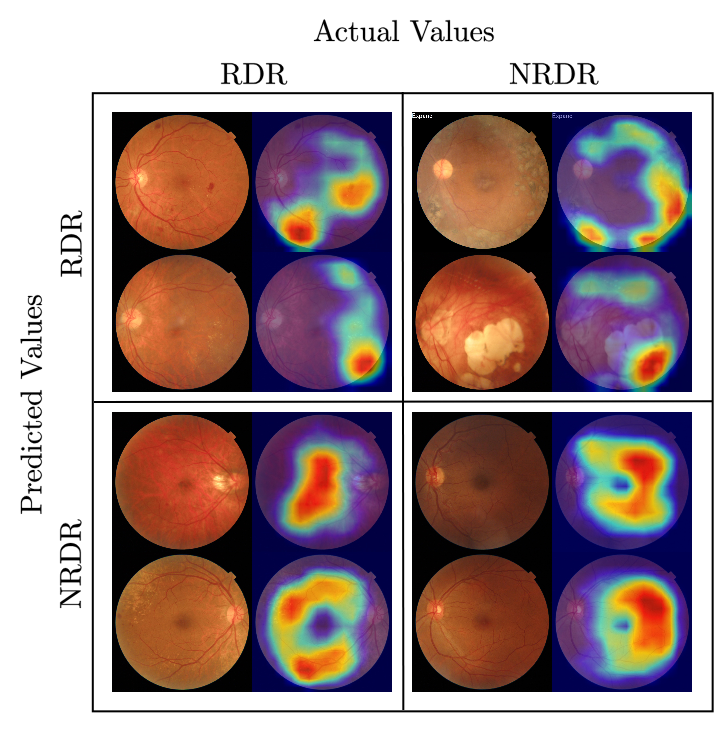}
    \caption{Confusion matrix and feature maps using GradCam~\cite{selvaraju2017grad}. It is possible to see the most important features of the model (M1-Referral) in each case. }
    \label{fig:expla}
\end{figure}

One advantage of creating a model is that the explanation can be performed. 
Figure \ref{fig:expla} presents a confusion matrix with image heatmaps. Two examples of True positive cases (True: RDR, Pred: RDR), show that the feature maps highlight the hemorrhages, microvascular abnormalities, and exudates that are typical characteristics of R3 (Severe non-proliferative DR). 
The false positive cases (True: NRDR, Pred: RDR) highlight LASER lesion areas, exudates, and anatomical deformations. In true negative cases (True:NRDR, Pred:NRDR), the heatmaps indicate that the model focuses mainly on the arcades and fovea area. False negative RFP cases (True:RDR, Pred:NRDR) occur when the model predicts R2, but the actual value is R3.

\subsection{Deploy, monitoring, and exposure to external review}
The RAIS-DR has been deployed at \url{https://gitlab.com/inteligencia-gubernamental-jalisco/jalisco-retinopathy} with a Responsible AI license. 
The system  utilizes a REST-API with the AI-based system using in Docker container, which is ready to use in the cloud serverless. 
The system allows the tester users to upload several images and view each model's (quality models and reference and DR-grade models) results (class and confidence interval). In addition,  expert users can suggest alternative quality levels and DR grades to monitor the RAIS-DR performance.

Regarding external review,  this project has been submitted at various stages to conferences~\cite{a_sanchez_lxai_2019, briceno2020automatic,bermejoetiquetado}, research papers~\cite{gonzalez2020artificial,pinedo2022suitability, cortes2021fair}, and international committees such as GPAI~\cite{gpai-2020} and UNESCO's Top-100 (\href{https://ircai.org/presenting-the-global-top-100-outstanding-projects-artificial-intelligence-based-referral-system-for-patients-with-diabetic-retinopathy-in-jalisco/}{ where it selected in the top-10}).  These practices increased the transparency of the RAIS-DR and yielded feedback.

\subsection{Model limitations}
The model cannot detect other retinal diseases, such as glaucoma or macular age degeneration. Moreover, image artifacts, low-quality images, LASER-scarring, and unusual anatomical changes in the retina will significantly reduce the confidence/performance of DR classification models.

\section{Conclusions}\label{sec:conlusions}

This study introduced a novel Responsible AI-based System for Diabetic Retinopathy Screening (RAIS-DR) in a clinical environment, using a separate local dataset comprising approximately 1,000 patients. The here presented approach is grounded in ethical principles and specific actions in the AI lifecycle. 
Validation in a clinical environment underscores the practicality and real-world application of this research, enhancing the typical AI lifecycle with responsible actions to mitigate \textit{bias} and assess risks effectively. 
Key features of  RAIS-DR include preprocessing techniques to prevent geometric distortions, reduce background noise, evaluate image quality during acquisition, and incorporate external evaluations throughout the AI cycle.

The RAIS-DR, was numerically compared with Eyeart, demonstrating superior specificity and effectively reducing false positives. It also exhibited a higher Positive Predictive Value (PPV) than Eyeart, indicating increased reliability when classifying instances as positive. 
The system outperformed  Eyeart, accurately identifying both positive and negative cases with greater precision. 
Although the proposed model showed lower sensitivity, this presents an opportunity to improve further reduce false negatives. The model's AUC curves indicate the potential to adjust thresholds to address the false-negative rate if necessary.
Additionally, fairness metrics confirm that RAIS-DR performs equitably across different subgroups. 
Using home made model RAIS-DR offers additional advantages, including customization and full access to the model, thus facilitating better understanding explainability and continuous improvements.   


\section*{Acknowledgments}
The authors gratefully acknowledge NVIDIA Corp., the Jalisco
Government, and the LXAI Supercomputing Program for providing access to the
DGX-A100. We also thank EyePACS, Messidor II, and IDRID for sharing their
datasets and annotations. Special appreciation goes to CONAHCYT, the GPAI
Scale-Up Solution Program for their valuable feedback and to the IDB Fair-LAC
Program for supporting local data collection.

\end{document}